\documentclass{article}

\usepackage[preprint, nonatbib]{main}

\usepackage{graphicx}
\usepackage{booktabs}
\usepackage{multirow}
\usepackage{multicol}
\usepackage{amsmath}
\usepackage{adjustbox}
\usepackage[format=plain,labelformat=simple,labelsep=period,font=small,compatibility=false]{caption}
\usepackage{wrapfig}
\usepackage{xcolor}

\usepackage[accsupp]{axessibility}  

\usepackage[pagebackref,breaklinks,colorlinks]{hyperref}
\usepackage[capitalize]{cleveref}
\crefname{section}{Sec.}{Secs.}
\Crefname{section}{Section}{Sections}
\Crefname{table}{Table}{Tables}
\crefname{table}{Tab.}{Tabs.}

\DeclareMathOperator*{\argmax}{arg\,max}

\newcommand\minisection[1]{\vspace{1mm}\noindent \textbf{#1}}
\newcommand{\model}{P$\text{R}^2$T-NeRF}

\usepackage[utf8]{inputenc} 
\usepackage[T1]{fontenc}    
\usepackage{hyperref}       
\usepackage{url}            
\usepackage{booktabs}       
\usepackage{amsfonts}       
\usepackage{nicefrac}       
\usepackage{microtype}      
\usepackage{xcolor}         

\title{\underline{P}oint \underline{R}esampling and \underline{R}ay \underline{T}ransformation \\ Aid to Editable NeRF Models}

\author{%
  Zhenyang Li\thanks{Equal contribution} \\
  The University of Hong Kong \\
  \And
  Zilong Chen$^{*}$ \\
  Tsinghua University \\
  \And
  Feifan Qu \\
  The University of Hong Kong \\
  \And
  Mingqing Wang \\
  Tsinghua University \\
  \And
  Yizhou Zhao \\
  Carnegie Mellon Uniersity \\
  \And
  Kai Zhang \\
  Tsinghua University \\
  \And
  Yifan Peng \\
  The University of Hong Kong \\
}

\begin{document}

\maketitle

\begin{abstract}
  In NeRF-aided editing tasks, object movement presents difficulties in supervision generation due to the introduction of variability in object positions.
  Moreover, the removal operations of certain scene objects often lead to empty regions, presenting challenges for NeRF models in inpainting them effectively.
  We propose an implicit ray transformation strategy, allowing for direct manipulation of the 3D object's pose by operating on the neural-point in NeRF rays.
  To address the challenge of inpainting potential empty regions, we present a plug-and-play inpainting module, dubbed \emph{differentiable neural-point resampling (DNR)}, which interpolates those regions in 3D space at the original ray locations within the implicit space, thereby facilitating object removal \& scene inpainting tasks. Importantly, employing DNR effectively narrows the gap between ground truth and predicted implicit features, potentially increasing the mutual information (MI) of the features across rays. Then, we leverage DNR and ray transformation to construct a point-based editable NeRF pipeline (\model).
  Results primarily evaluated on 3D object removal \& inpainting tasks indicate that our pipeline achieves state-of-the-art performance. In addition, our pipeline supports high-quality rendering visualization for diverse editing operations without necessitating extra supervision. Additional results are available in the \href{https://sample-nerf.github.io/}{Demos}.
  
\end{abstract}

\section{Introduction}

The pursuit of achieving full flexibility in manipulating scene representations is a prominent objective in vision and graphics communities. The ability to manipulate various aspects of a scene, such as its location or shape, while achieving visually stunning results efficiently, is in high demand~\cite{liu2023nero,srinivasan2021nerv}. However, challenges pertaining to the consistency and authenticity of the synthesis persist. Apart from the deformation and morphing operations extensively discussed in state-of-the-arts~\cite{chen2023neuraleditor,NeRFshop23}, operations such as scene object removal \& inpainting, as well as location transformation, are essential in scene editing applications.

Recent advancements in editable 3D reconstruction and rendering primarily build upon Neural Radiance Fields (NeRFs)~\cite{original.nerf,nerf.editing,nerf.in,mirzaei2023spin,yin2023ornerf}, which leverage well-trained 2D inpainting models. Most research efforts have focused on constructing robust supervision mechanisms and developing intricate network architectures to enhance the editing capabilities. However, considering the objective of the editing task, the removal operations, which depend directly on manipulating rays and points~\cite{xu2022deforming-nerf,peng2022cage-nerf,tseng2022cla-nerf,garbin2022voltemorph}, have received little attention. Meanwhile, exploring an operation capable of augmenting the prior knowledge of NeRF rendering can be highly valuable, as it can facilitate the convergence and guide the higher-fidelity reconstruction of backgrounds.

It is worthy noted that NeRF representation can be regarded as an advanced version of MPI (Multi-Plane Image) \cite{tucker2020single} due to its capability to model the scene with dense planes, which contains differentiable and continuous features.
This insight inspires us to investigate an interpretable and generalizable strategy for constructing prior features before using NeRF to render empty regions, along with flexible editing operations for object transformation, including removal. 

To be more specific, the editing process can be divided into two steps: object editing and empty regions inpainting. To support flexible editing effects, we explore directly manipulating rays that correspond to the specified object and defining the general transformation as rigid and non-rigid types. Next, we derive corresponding transformations in neural point cloud and NeRF rays, including rotation and translation. For the detailed removal process, we extract a 2D mask from the single unedited annotated source image, and then unproject this mask onto a 3D point cloud with rasterization and registration for target segmentation. For rotation, translation, and scaling operations, we directly manipulate the NeRF rays corresponding to the 3D mask projected from the 2D mask to manipulate the pose and shape of the object.

Notably, the edited point cloud leaves \textit{empty spaces} in the masked regions, potentially leading to destruction of existing features during the subsequent training and fine-tuning process.
To tackle this issue, we dive into general editing tasks with respect to the information theory and prove that these aggregate strategies can increase the \emph{mutual information} (MI)~\cite{wang2022rethinking} between rays, which aids the task performance. We further implement the Differentiable Neural-Point Resampling (DNR) strategies for inpainting initialization, which interpolate the empty locations with the surrounding implicit features. The strategies aim to replace the non-differentiable aggregate operations in Point-NeRF \cite{xu2022point} or NeuralEditor \cite{chen2023neuraleditor}.
Our technical contributions are as follows:
\begin{itemize}
    \item We propose a plug-and-play, differentiable inpainting  DNR scheme for feature aggregation and validate its effectiveness with the information theory.
    \vspace{1pt}
    \item We derive the general formulation of scene editing in point-based NeRFs, as well as explore the robust removal, rotation, and translation of implicit features and rays.
    \vspace{1pt}
    \item We construct an editable NeRF pipeline which delivers state-of-the-art results on scene removal \& inpainting benchmarks, including extensive video evaluations (\href{https://sample-nerf.github.io/}{Demos}) of novel view synthesis under different processing settings, with a significantly drop of training time.
\end{itemize}


\section{Related Work}
\label{sec:related work}
\vspace{-0.2cm}
\paragraph{3D Neural Representation:}
\label{subsec:nerfs}
Conventional 3D representations, including explicit data formulation such as meshes \cite{mesh.wang2018pixel2mesh}, point clouds \cite{wang2019mvpnet,achlioptas2018learning}, volumes \cite{qi2016volumetric}, and implicit functions \cite{mescheder2019occupancy,niemeyer2020differentiable,yariv2020multiview}, have been intensively studied in computer vision and graphics applications.
Recently, NeRF~\cite{original.nerf} has emerged as a significant breakthrough in 3D scene reconstruction. Consequently, these aforementioned representations have been transferred into neural technologies, leading to state-of-the-art performance.
While vanilla NeRFs have demonstrated remarkable progress, subsequent works have primarily focused on addressing their limitations. These efforts aim to improve reconstruction quality \cite{barron2021mip,guo2022nerfren,verbin2022ref}, inference efficiency \cite{yu2021plenoctrees,wu2022diver,fridovich2022plenoxels}, and cross scenes or style generalization \cite{chen2021mvsnerf,xu2022point,yu2021pixelnerf}, yielding numerous promising results. However, it is crucial to also prioritize research in object and scene editing, which we delve into in this work.

\paragraph{Point Cloud-based 3D Reconstruction:}
\label{subsec:3D Reconstruction via pc}
Point clouds have emerged as a flexible representation that offers several advantages, including low computation and storage cost, and ease of collection for accurate depth estimation. Recently, point clouds have gained popularity for rendering surfaces from 3D to 2D images \cite{lassner2021pulsar,wiles2020synsin}.
Point clouds representation, however, has some apparent defects, such as empty regions and outlier points, that have gradually been overcome with the integration of neural rendering methods \cite{aliev2020neural,kopanas2021point,meshry2019neural}.
Researchers have addressed this issue by rasterizing neural features extracted from images, such as storing the features in the point cloud, as in Point-NeRF \cite{xu2022point}, which utilizes 3D volume rendering and achieves good performance. Nevertheless, Point-NeRF solely focuses on point locations in reconstruction and disregards the specific characteristics of point clouds.

\paragraph{NeRF-based Object and Scene Manipulation:} \label{subsec:NeRF-based Manipulation}
NeRF-empowered editing tasks have been well studied recently, leading to numerous promising methods and models~\cite{yuan2022nerf,wang2022clip,chen2023neuraleditor,nerf.in,kobayashi2022decomposing}. To name a few, Object-Compositional NeRF \cite{yang2021object-nerf} enables high-level scene/object adding/moving manipulation. Editing-NeRF \cite{liu2021editing} presents a conditional NeRF model that allows users to interact with 3D shape scribbles at the category level. 
Additionally, Removing-NeRF \cite{weder2023removing} utilizes a sequence of RGB-D images to facilitate the removal process and is available for users to specify masks.

However, these methods heavily rely on the supreme inpainting and segmentation supervision, hindering their generative capabilities. In this context, \model differs in that we support editing the scene directly on both NeRF rays and point clouds, thereby potentially leading to a unified framework that represents rotation, translation, scaling, as well as removal \& inpainting.

\section{Method}
\begin{figure}[t]
  \centering
  \includegraphics[width=1.0\linewidth]{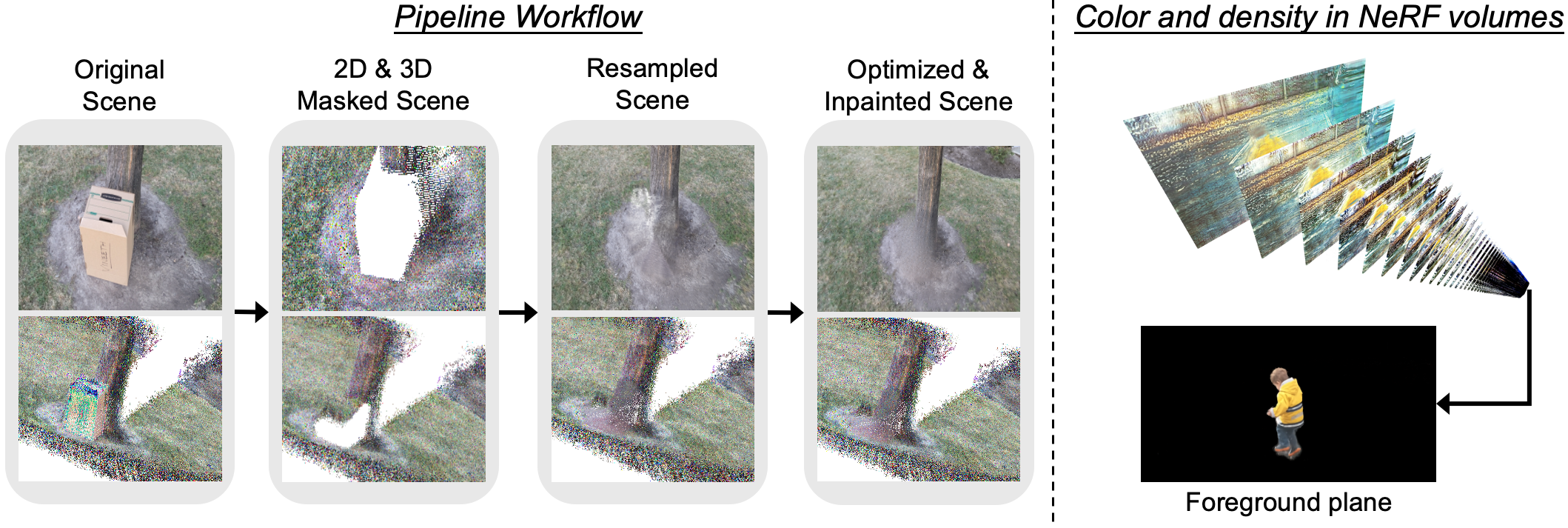}
  \captionof{figure}{\textbf{\model supports robust manipulations for novel view synthesis and scene editing}, in particular object removal \& inpainting. Left: Workflow and progressive results of our pipeline in object editing of a natural scene; Right: Distribution of color (features) and density in the volume grid of vanilla NeRF, where the example scene~\cite{li2021neural} is decomposed into quasi-continuous planes alone depth while object features spread out with varying degrees on different planes. We observe that the foreground only shows a kid within shallow depth (planes in black), while the far-depth background gradually appears in the planes behind.}
  \label{fig:teaser}
\end{figure}
\subsection{Motivation and Preliminary}\label{subsec:moti-exps}
In NeRF volume, the color with its distribution of density features across multiple planes, as illustrated in \cref{fig:teaser} (right), reveals a noteworthy pattern that becomes more pronounced with increasing depth. More specifically, pixels situated at different depth locations within the scene exhibit distinct characteristics, wherein foreground objects predominantly exhibit features distributed on planes with shallow depths, while objects closer to the background showcase features distributed on planes with greater depths. These observation effectively illustrates the variation in rays traversing through both the foreground character and the background, implying the potential for achieving object editing by directly manipulating the rays associated with the specific object.

\subsection{Editing with NeRF} \label{subsec:location-edit}
\paragraph{General Problem Formulation.}
Given the rays sampled by a NeRF as a set $R=\{ r_i \}_{i=1}^{N}$, where $N$ refers to the number of rays, for each ray, NeRF samples $N_i$ points in the range of $t_n$ and $t_f$ (near and far depth). The $j^{th}$ point on the $i^{th}$ ray is denoted as $p_{ij}$. Here, we can define the general implicit editing operations in rigid and non-rigid, separately.
The general scene editing operations function $\mathcal{F}$, and then we can manipulate the ray $r_i$ with its neural points set $\{p_{ij}\}_{j=1}^{N_i}$:
\vspace{-0.3cm}
\begin{align} \label{eq:general edit}
    r_i' = \mathcal{F} \odot r_i \,\, \vee \,\, p_{ij}' = \mathcal{F} \odot p_{ij},
\end{align}
where, $r_i'$ and $p_{ij}'$ are the edited rays and points respectively, and $\odot$ is a general math operator, representing Rigid and Non-rigid transformation.

\begin{figure*}[t]
    \includegraphics[width=1.0\linewidth,height=5.5cm]{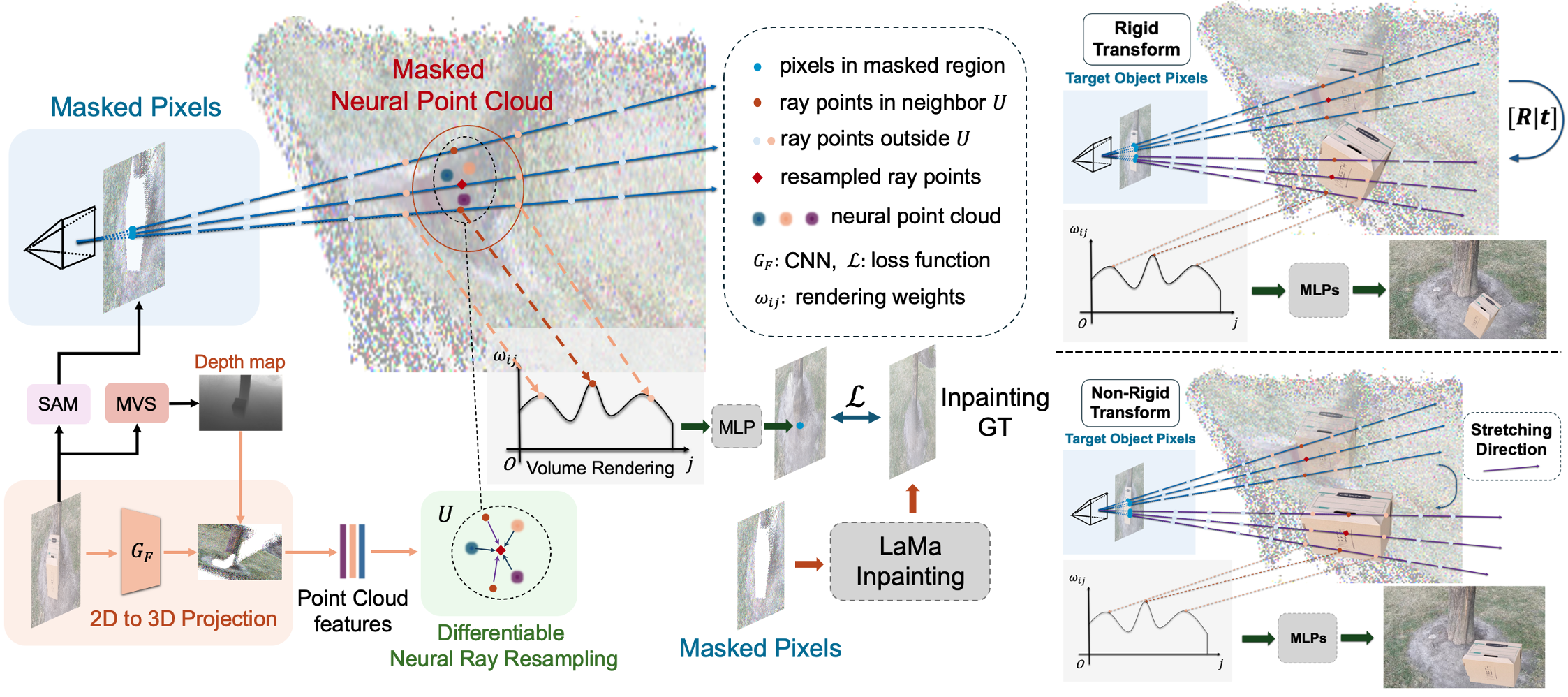}
    \caption{\textbf{Overview of our editable rendering pipeline and transformations.} Left: The object removal \& inpainting framework integrates the SAM model to generate a 2D mask for the target object. Subsequently, this 2D mask is unprojected onto the 3D space, effectively creating a point cloud mask, while features are extracted from the original image to serve as point cloud features and fine-tuning to derive the neural point cloud of the unedited scene. Next, the DNR module is utilized to mend the features of empty regions in masked 3D points. Finally, we supervise the rendering views by generating inpainted images from LaMa; Right: Schematic visualization of rigid and non-rigid transformations.}
    \label{fig:remove-pipeline}
\end{figure*}

\vspace{2pt}
a) \underline{Rigid Transformation of Implicit Rays.}\label{para:rigid}
When operating the rigid transformation to the target rays with the rotation matrix $\textbf{R} \in \mathbb{R}^{3 \times 3}$ and the translation matrix $\textbf{t} \in \mathbb{R}^{3 \times 1}$, the rigid transformation matrix is formulated as $\textbf{T}_{rigid}=\left(\begin{smallmatrix} \textbf{R} & \textbf{t} \\\textbf{0} & 1 \end{smallmatrix} \right)$. Here, we have the ray $r_i=(o_i, d_i)$ in the world coordinate with its origin $o_i \in \mathbb{R}^{3 \times 1}$ and direction $d_i \in \mathbb{R}^{3\times 1}$, so to obtain the transformed ray $r_i'=\textbf{T}_{rigid}\left(\begin{smallmatrix} o_i^\text{T} & 1 \\ d_i^\text{T} & 1 \end{smallmatrix}\right)^\text{T} $. 
The corresponding point $p_{ij} \in \mathbb{R}^{3 \times 1}$ on the ray is transformed in the same scheme, i.e.,  $p_{ij}'=\textbf{T}_{rigid}\left(\begin{smallmatrix} p_{ij}^\text{T}, & 1 \end{smallmatrix}\right)^\text{T}$. Notably, for rigid transformation, ray transformation is constantly equivalent to points transformation (\cref{fig:remove-pipeline} top-right).

\vspace{2pt}
b) \underline{Deformable Transformation of Implicit Rays.}\label{para:non-rigid}
Non-rigid transformation includes scaling and shearing, which the offset varies for each point, and is not equivalent to ray transformation. Thus, we specify the edited ray $r_i'$ by collecting the edited points $p_{ij}'$ as a new set.
Concerning the point offset on $p_{ij}$, we denote $\delta p_{ij}=(\delta p_{ijx}, \delta p_{ijy})$ and express $p_{ij}'=p_{ij}+\delta p_{ij}$, consequently, yielding a new set of points for the transformed ray $r_i'$, represented as $\{ p_{ij}' \}_{j=1}^{N_i}$ (\cref{fig:remove-pipeline} bottom-right).
\vspace{-0.3cm}
\paragraph{Target Implicit Segmentation.}\label{subsec:edit-tar}
After applying general transformation ($\odot$) to the masked regions, we aim to produce the target 3D point cloud segmentation mask. Our approach, akin to OR-NeRF \cite{yin2023ornerf} and SPIn-NeRF \cite{mirzaei2023spin}, leverages the masks obtained from the multi-view images for scene removal tasks, to generate the mask with the following steps, as shown in \cref{fig:remove-pipeline} (left).

\vspace{2pt}
a) \underline{Point Cloud Initialization.}
To initialize the point cloud of the unedited image, we employ TransMVSNet \cite{ding2022transmvsnet}, a state-of-the-art cost volume based depth estimation model, to predict the depth for each pixel.
The final depth map $D \in \mathcal{R}^{H \times W}$ can be regressed in weighted linear sum using the depth in planes and the corresponding probabilities, where $H$ and $W$ represent the height and width of the image resolution, respectively. We then regard the depth as the true location in 3D space of the pixels and unproject them to obtain the point cloud $\mathcal{P}=\{(x_i, y_i, z_i)\}_{i=1}^{N_P}$.

\vspace{3pt}
b) \underline{2D Inpainting Mask Configuration.}
Given the single view $I$ with its camera extrinsics $[\text{R}|\text{t}]$ and intrinsics $\text{K}$, along with the corresponding predicted neural point cloud $\mathcal{P}=\{ (x_i, y_i, z_i, f_i, \omega_i) \}_{i=1}^{N_P}$, a CNN encodes image $I$ into semantic features associated with 2D locations, which are then projected onto the corresponding 3D points as $\{ f_i \}_{i=1}^{N_P}$ using rasterization. Herein, $\omega_i \in [0,1]$ represents the estimation accuracy of 3D locations. For a better comparison with OR-NeRF, we employ the same set of point prompts $Pr=\{ (u_i, v_i) \}_{i=1}^{N_{Pr}}$ and utilize the segmentation model SAM \cite{kirillov2023segment} denoted as $S$ to remove the target object. Consequently, we can represent the mask $M$ with the include pixel coordinates conditioned on $Pr$ as $M=S(I|Pr)=\{ (u_i, v_i)\}_{i=1}^{N_{M}}$, where $N_{M}$ is the number of pixels involve in $M$. Since our datasets are not used in the training process of SAM, the segmentation boundaries are not sharp enough. We resolve the above depth and RGB inconsistency by the constraints in \cref{subsec:optimize}, and optimize the 3D implicit features (\cref{subsec:diff-ray}).

\vspace{2pt}
c) \underline{3D Neural Points Mask Configuration.}
Based on the camera parameters $[\text{R}|\text{T}]$, $\text{K}$, and the coordinates of the mask region $M$, we compute the set of 3D masked points $\mathcal{\hat{P}}=(\text{K}[\text{R}|\text{t}])^{-1}(DM)$. The points in $\mathcal{\hat{P}}$ should align with the partially estimated point cloud $\mathcal{P}$, and the resulting registered point cloud is denoted as $\mathcal{\hat{P}}_{NN}$. Ultimately, we obtain the point cloud $\mathcal{P}_M= \mathcal{P} \setminus \mathcal{\hat{P}}_{NN}$, which excludes points locate on the target object.

\subsection{Differentiable Neural-Point Resampling (DNR)} 
\paragraph{Information Entropy Analysis.} \label{subsec:infor-entro}
Drawing insights from zero-shot \cite{wang2022rethinking} and action recognition \cite{zhao2022alignment} tasks, it becomes evident that augmenting the mutual information derived from input data can contribute to improved task performance. In this section, we present a theoretical exposition demonstrating that feature aggregation strategy elevate the mutual information ($\text{MI}$) among rays.

\vspace{2pt}
a) \underline{Problem Definition.} \label{para: prob def}
$\text{MI}$ quantifies the shared information between two features, reflecting the degree of similarity between them. Consequently, a higher degree of similarity implies a stronger indication of their interrelatedness. Given the ray $r_i$ with one of its top-K nearest rays $r_j$, while $f(r_i)$ represents the features on $r_i$. The objective for the aggregation problem is defined as
\begin{align} \label{eq:prob def}
    \kappa (f(r_i), f(r_j)) = \argmax_\alpha \text{MI} [\alpha(f(r_i)), f(r_j)].
\end{align}
In this context, the symbol $\alpha$ represents a specific operation of $\kappa$ that aggregates features from the $r_j$ to $r_i$. In the inpainting task, the ray $r_i$ resides within the masked region where features have been removed (\cref{subsec:edit-tar}), while $r_j$ corresponds to a ray in the unmasked area. Consequently, $r_i$ exhibits no discernible relationship with $r_j$, resulting in a low $\text{MI}$. \cref{eq:prob def} is designed to enhance $\text{MI}$ through the utilization of the operation $\alpha$, which effectively transfers partial information from $r_j$ to $r_i$.

\vspace{2pt}
b) \underline{Proof and Analysis.}
Recall the objective in \cref{eq:prob def}, we simplify $f(r_i)$ and $f(r_j)$ as $f_i$ and $f_j$, and $\kappa$ can be expressed as
\begin{align}\label{eq:obj_mi}
    \kappa(f_i,f_j)=\argmax_\alpha \text{MI}[\alpha(f_i), f_j].
\end{align}
In the condition that the rays obey the same distribution in the same scene, we define the $\text{MI}$ as
\begin{align}\label{eq:mi}
    \text{MI}[\alpha(f_i),f_j]=\text{H}(\alpha(f_i))-\text{H}(\alpha(f_i)|f_j).
\end{align}
The information entropy, denoted as $\text{H}$, characterizes the likelihood of an event occurring. In NeRF models, this is construed as the presence of a specific feature on the ray, represented as light transmittance. The features of sampled points on rays $f_i$ and $f_j$ are expressed as sets $f_i=\{ f_{ik} \}_{k=1}^{N_i}$ and $f_j=\{ f_{jk} \}_{k=1}^{N_j}$, respectively. In accordance with the mathematical definition of $\text{H}$ and NeRF models, we have
\begin{align} \label{eq:infor-entro}
    \begin{split}
        \text{H}(\alpha(f_i))&=-\sum_{k=1}^{N_i} \text{P}(\alpha(f_{ik})) \log \text{P}(\alpha(f_{ik})), \\
        \text{H}(\alpha(f_i)|f_j)&=-\sum_{k=1}^{N_i} \text{P}(\alpha(f_{ik})|f_j) \log \text{P}(\alpha(f_{ik})|f_j).
    \end{split}
\end{align}
Observing that the operation $\alpha$ does not alter the point locations along the ray, the probability $\text{P}(\alpha(f_{ik}))$ remains equivalent to $\text{P}(f_{ik})$, representing the light transmittance before the applying $\alpha$. Consequently, $\text{H}(\alpha(f_i))=\text{H}(f_i)$. Regarding the conditional probability in \cref{eq:infor-entro}, with applying the Bayes' theorem, we can deduce the following
\begin{align}
    \text{P}(\alpha(f_{ik})|f_j)=\bigg[ \frac{\text{P}(\alpha(f_{ik}))}{\text{P}(f_j)} \bigg] \text{P}(f_j|\alpha(f_{ik})),
\end{align}
and for the condition without $\alpha$, $\text{P}(f_j|f_{ik})$ is tending to 0, leading to a greater absolute value of $\text{P}(f_j|\alpha(f_{ik}))$. We have $\text{H}(f_i|f_j)$ as follows
\begin{align} \label{eq:before dnr hij}
    \text{H}(f_i|f_j)=-\sum_{k=1}^{N_i} \text{P}(f_{ik}|f_j) \log \text{P}(f_{ik}|f_j).
\end{align}
Upon comparing the conditional entropy in \cref{eq:infor-entro} with that in \cref{eq:before dnr hij}, we observe that $\text{H}(\alpha(f_i)|f_j) \leq \text{H}(f_i|f_j).$
Given \cref{eq:mi}, and $\text{MI}[f_i,f_j]=\text{H}(f_i)-\text{H}(f_i|f_j)$,
We derive that $\text{MI}[\alpha(f_i),f_j] \geq \text{MI}[f_i,f_j]$,
which the aggregation operation $\alpha$ optimizes the objective function of \cref{eq:obj_mi}. In a broader context, optimizing \cref{eq:obj_mi} serves to augment the MI among rays. This augmentation leads to improved model performance, particularly in the inpainting task.

\begin{figure}[t]
    \centering
    \includegraphics[width=0.8\linewidth]{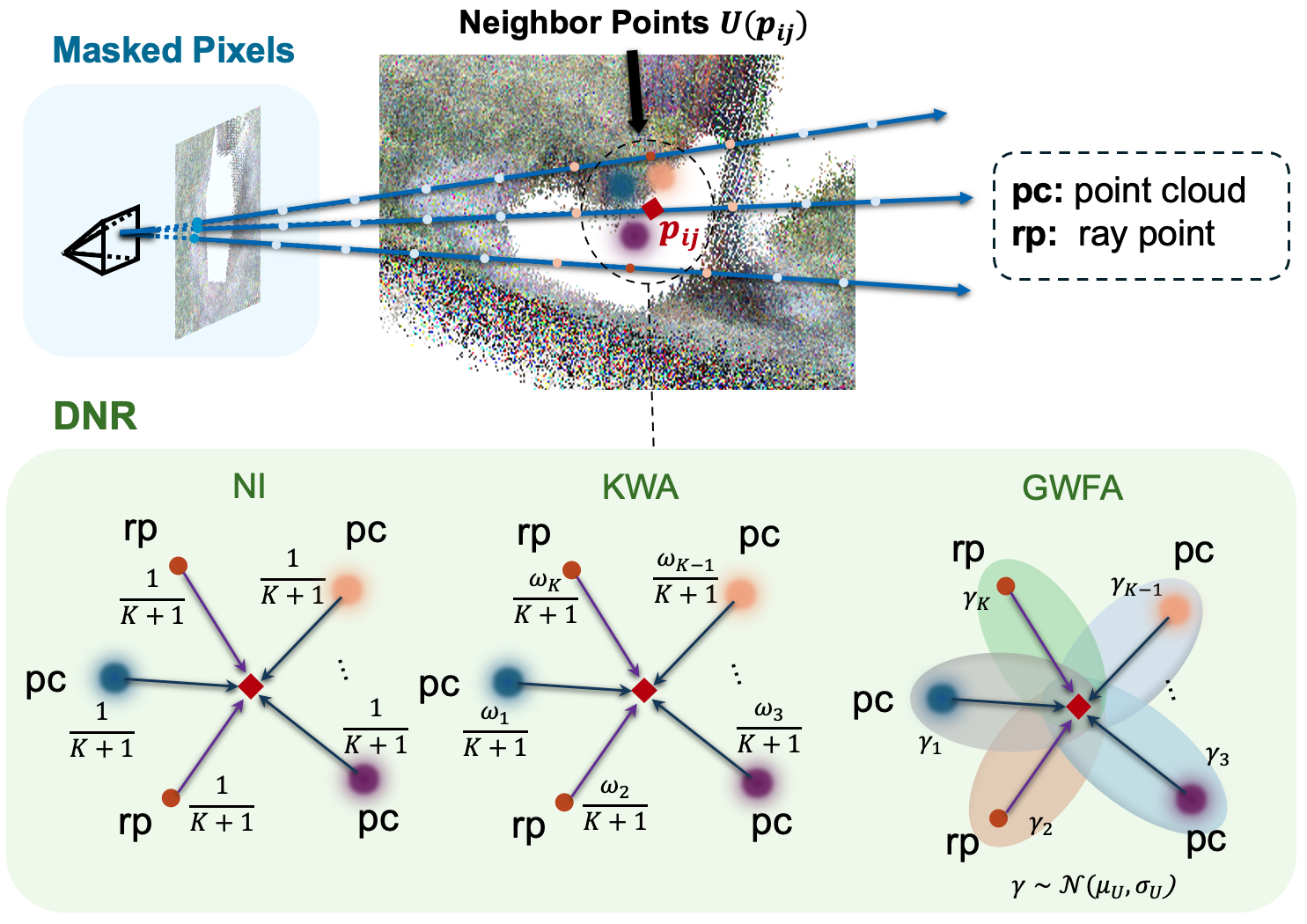}
    \caption{\textbf{Overview of DNR strategies in \cref{subsec:diff-ray} c).} The three feature resampling schemes (NI, KWA, and GWFA) are illustrated, with different sum weights defined.}
    \label{fig:overview-dnr}
\end{figure}

\vspace{2pt}
c) \underline{Problem Solution - DNR.}\label{subsec:diff-ray} 
The crucial task to address the issue outlined in a) is the operation $\alpha$. In this regard, we aim to aggregate neighboring features to the rays within the masked regions and maximize the objective (\cref{eq:obj_mi}). As depicted in \cref{fig:overview-dnr}, we visualize the proposed DNR strategies.
Given the feature $f_{ij}$ associated with the $j^{th}$ point $p_{ij}$ on ray $r_i$, derived from the pretrained model in \cite{xu2022point}, we aggregate neighbor features onto this existing feature $f_{ij}$, by defining the top-$K$ neighbor points set for the point $p_{ij}$ on the ray as $U(p_{ij})=\{ p_k \}_{k=1}^{K}$, where the feature of $p_k$ is denoted as $f(p_k)$. Below are three versions of DNR implementation, each buildings upon the previous one in a progressive manner.

\vspace{6pt}
\textbf{\romannumeral1.} Neighbors Interpolation (NI):
\label{subsubsec:nli}
We first implement the resampling module NI, which interpolates the features of point clouds in masked regions using the surrounding neural points. The purpose of NI is to evaluate the effectiveness of resampling methods. NI refers to the geometric average of the features in $U(f(p_{ij}))=\{ f(p_k) \}_{k=1}^{K}$, then we have the updated feature $f(p_{ij})'$ as
\begin{align} \label{eq:ni}
    f(p_{ij})'=\frac{1}{K+1} \bigg[f(p_{ij})+\sum_{k=1}^K f(p_k) \bigg].
\end{align}
Equation (\ref{eq:ni}) represents a straightforward yet effective method for performing inpainting operations in implicit space. Additionally, we note that within this module an empty 3D location may possess similar feature and geometry information, attributed to the inherent space consistency.

\vspace{6pt}
\textbf{\romannumeral2.} KNN Weighted Average Resampling (KWA):
\label{subsubsec:kwa}
Next, given that the confidence of the features matters in the aggregation process, we have the confidence $\omega_k$ of the feature $f(p_k)$ as the confidence weight
\begin{align} \label{eq:kwa}
    f(p_{ij})'=\frac{1}{K+1} \bigg[ \bigg(1 - \frac{1}{K} \sum_{k=1}^K \omega_k \bigg) f(p_{ij}) + \sum_{k=1}^K \omega_k f(p_k) \bigg] .
\end{align}
The original feature in the pretrained model, denoted as $f(p_{ij})$, undergoes a weighting process as indicated by the equation above. For the point $p_{ij}$, it retains the original feature if the features within $U(p_{ij})$ exhibit a low average confidence; otherwise, it places greater reliance on the surrounding features.

\vspace{6pt}
\textbf{\romannumeral3.} 3D Gaussian Weighted Feature Aggregation (GWFA):
Despite of the confidence of each point feature, the impact (e.g. Similarity) of the surrounding features toward the center point is another crucial measurement, as $\gamma$. As for a 3D point, the impact of feature increases when a neighbor point gets closer to it, while it will decrease rapidly when the distance between them gradually becomes large. Assuming that $\gamma \sim \mathcal{N}(\mu, \sigma)$, we have
\begin{align}
    \begin{split}
        &f(p_{ij})'=\frac{1}{\sum_{k=1}^K \gamma(p_k)} \sum_{k=1}^K \gamma(p_k) f(p_k) ,\\
        \text{where} \quad &\gamma(p_{k})=\frac{1}{\sqrt{2\pi}\sigma_U}\exp{\bigg( -\frac{\|f(p_k)-\mu_U\|^2}{2\sigma_U^2} \bigg)},
    \end{split}
\end{align}

where $\mu_U = \frac{1}{K}\sum_{k=1}^K f(p_k)$ and $\sigma_U^2 = \frac{1}{K}\sum_{k=1}^K \| f(p_k) - \mu_U \|^2$, both obeying the Gaussian distribution.
\vspace{-0.3cm}
\subsection{Optimizing Pipeline in Removal \& Inpainting}\label{subsec:optimize}
Figure \ref{fig:remove-pipeline}(a) shows the pipeline for the object removal \& inpainting task. In the lack of supervision for mask and inpainting in the current dataset, we employ the well-trained SAM \cite{kirillov2023segment} to generate the mask image corresponding to the target object. Subsequently, for the masked image, we employ the LaMa \cite{lama} model to inpaint the masked region, consequently enabling the inpainting supervision. Overall, the training of \model involves the following weighted loss terms
\begin{equation}\label{eq:total ft loss}
    \mathcal{L}=\mathcal{L}_{color} + \lambda_{per} \mathcal{L}_{per} + \lambda_{depth} \mathcal{L}_{depth} + \alpha \mathcal{L}_{sparse} ,
\end{equation}
where $\mathcal{L}_{color}$ is the RGB reconstruction loss for unsegmented pixels outside the SAM mask regions, $\mathcal{L}_{per}$ is the perceptual loss LPIPS \cite{zhao2021large}, and $\mathcal{L}_{depth}$ calculates the $\ell_2$ distance between the prediction and ground truth depth. Additionally, we introduce a sparse loss~\cite{xu2022point} from Point-NeRF upon the point confidence.
\vspace{-0.2cm}
\paragraph{Per-scene Fine-tuning.}
To generate the original point cloud features, we initially train NeRF solely with $\mathcal{L}_{color}$, supervised by unedited mono-view scene data.
Then, leveraging the LaMa pretrained model, we generate RGB and depth inpainting ground truth for the masked regions in each scene, and supervise the inpainting outcomes from our model, resulting in high-quality rendered images. Optimization of the inpainting outcomes for each scene is conducted independently,  and we focus primarily on the perceptual loss on the masked area.

\section{Implementation and Results}\label{sec:implement}
\subsection{Preliminaries and Configurations}
\paragraph{Point Cloud Initialization.} The accuracy of a point cloud heavily relies on the accuracy of depth estimation. The original MVSNet \cite{yao2018mvsnet} often introduces significant noise and blur when estimating the depth map, resulting in fuzzy prediction, particularly on large scene datasets. 
Instead, we introduce TransMVSNet \cite{ding2022transmvsnet} for depth estimation, which produces sharper and more precise object boundaries, and then project the pixels to point cloud using the known extrinsic, intrinsic parameters of the camera and depth map.
\begin{figure*}[t]
  \centering
    \includegraphics[width=1.0\linewidth]{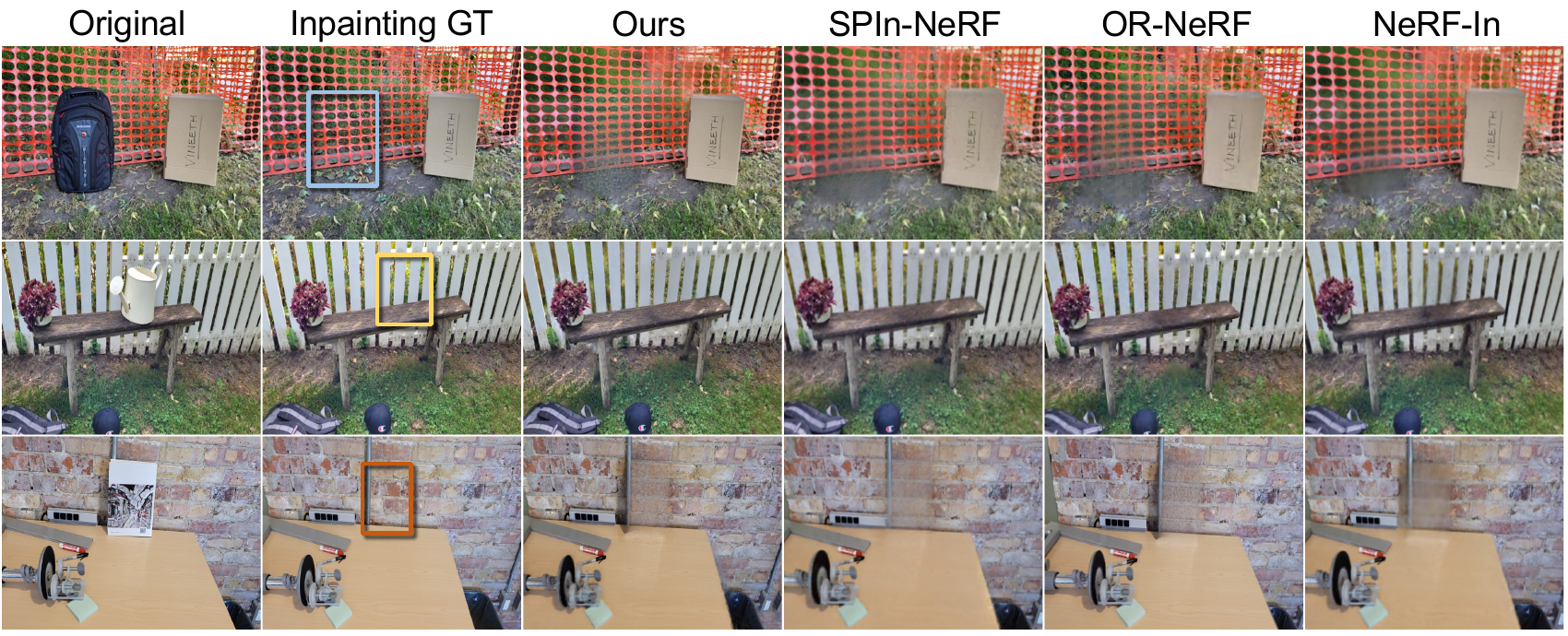}
    \caption{\textbf{Qualitative comparison of \model with counterparts.} A comparative analysis of inpainting results is conducted across three scenes using the SPIn-NeRF dataset \cite{mirzaei2023spin}. The color frames in the ``Inpainting GT'' column indicate the locations of the target object to be removed. Columns (From $4^{\text{th}}$ to $6^{\text{th}}$): the novel view of the scene generated by Ours, SPIn-NeRF, OR-NeRF, and NeRF-In. It should be noted that the recovery of shadows and periodic textures proves challenging both for baselines and our model, nevertheless, our model demonstrates superior performance in alleviating the shadows and twisted artifacts in textures in the rendering results.}
   \label{fig:vis spin sota}
\end{figure*}
\paragraph{Benchmarks.}
Due to the absence of ground truth in the reconstruction datasets, we incorporate the SPIn-NeRF \cite{mirzaei2023spin} dataset as one of the benchmark, which includes human annotated object masks and captures of the scene after object removal. To ensure a diverse layout of the objects, we select 8 scenes from the SPIn-NeRF dataset, excluding two duplicate scenes. Additionally, we evaluate our pipeline on real objects data \cite{weder2023removing}. This set includes 16 scenes, each containing a single object of interest with varying background textures, object scales, and complex scene structures, thereby presenting additional challenges.
Further, we select 3 scenes from the commonly used 3D reconstruction dataset from IBRNet~\cite{wang2021ibrnet}. For further visualization, please refer to the Supplementary materials.
\vspace{-0.3cm}
\paragraph{Metrics.}
Concerning the removal and inpainting task, we adopt the same settings as in SPIn-NeRF \cite{mirzaei2023spin} and evaluate the results using the learned perceputual image patch similarity (LPIPS) \cite{zhao2021large}, the average Fréchet inception distance (FID) \cite{fid} between the distribution of inpainting results and the ground truth, as well as the PSNR \cite{hore2010image}. We also compare the results with Remove-NeRF with SSIM \cite{wang2004image} metric instead of FID. We report the average scores of the four metrics in all scenes.

\paragraph{Implementation Details.}
In order to resolve enhanced image features and depth estimation, we initially pretrain our pipeline on the DTU \cite{jensen2014large} dataset, which is commonly used for novel view synthesis.
After integrating the DNR modules into the pipeline, we conduct fine-tuning of \model on each scene with the objective of guiding the pipeline to learn from the removal and inpainting supervisions, which are generated from pretrained SAM and LaMa models.
Regarding the loss objective in \cref{eq:total ft loss}, we set ($\lambda_{per}$, $\lambda_{depth}$, $\alpha$) to (1e-2, 1e-3, 1e-4).

\subsection{Experimental Results} \label{sec:results}
\begin{table*}[t]
  \centering
  \caption{\textbf{Experiment results on scene object removal.} The first row indicates the scene object removal \& inpainting task datasets \cite{mirzaei2023spin,weder2023removing}. The full version of \model includes the perceptual loss and DNR module. Compared to other editable novel view synthesis works (OR-NeRF \cite{yin2023ornerf}, Remove-NeRF \cite{weder2023removing}) with multi-view input. Ours achieves the state-of-the-art results upon mono-view input. The annotation geo. refers to geometry guidance. No PSNR of NeRF-In \cite{nerf.in} on SPIn-NeRF dataset is available.}
  \adjustbox{width=1.0\linewidth}
  {\begin{tabular}{l|cc|c|c|cc||l|c|ccc}
    \hline
    ~ & \multicolumn{6}{c||}{SPIn-NeRF data \cite{mirzaei2023spin}} & ~ & \multicolumn{4}{c}{Real Objects data \cite{weder2023removing}} \\
    \hline
    ~ & \multicolumn{2}{c|}{Ours} & OR-NeRF & SPIn-NeRF & \multicolumn{2}{c||}{NeRF-In} & ~ & \multicolumn{1}{c|}{Ours} & \multicolumn{3}{c}{Remove-NeRF(Masked)} \\
    \hline
    ~ & w/o DNR & Full & Best & w geo. & Single & Origin & ~ & Full & Best & w depth & w/o depth \\
    \hline
    PSNR$\uparrow$ & 20.25 & \textbf{20.44} & 14.16 & 14.85 & - & - & PSNR$\uparrow$ & 24.81 & \textbf{25.27} & 25.01 & 24.23 \\
    FID$\downarrow$ & 50.92 & \textbf{50.17} & 58.15 & 156.64 & 183.23 & 238.33 & SSIM$\uparrow$ & 0.846 & \textbf{0.859} & 0.856 & 0.848 \\
    LPIPS$\downarrow$ & 0.401 & \textbf{0.330} & 0.676 & 0.465 & 0.488 & 0.570 & LPIPS$\downarrow$ & \textbf{0.096} & 0.125 & 0.128 & 0.130 \\
    \hline
  \end{tabular}}
  \label{tab:sota}
\end{table*}
\minisection{Mono-view Removal \& Inpainting.} 
Table \ref{tab:sota} presents a comprehensive comparison of the removal \& inpainting performance with other baselines. Our method consistently outperforms both 2D and 3D inpainting approaches across various evaluation metrics. Multiple versions of these methods are implemented in our comparison, clearly demonstrating the superiority of \model. For the SPIn-NeRF data, we achieve approximately 6 points higher PSNR, 8 points lower FID, and 0.12 points lower LPIPS compared to the second-best performing method. 
Concerning real-object data, we directly use the coarse masks provided by the dataset for the sake of comparison convenience. our method achieves a substantial improvement of 0.029 in LPIPS over Remove-NeRF \cite{weder2023removing}, albeit with a slight decrease in PSNR and SSIM. This discrepancy can be attributed to Remove-NeRF's incorporation of a view selection module, aimed at excluding incorrect inpainting supervision and addressing inconsistencies in views.

\begin{figure}[t]
    \centering
    \includegraphics[width=1.0\linewidth]{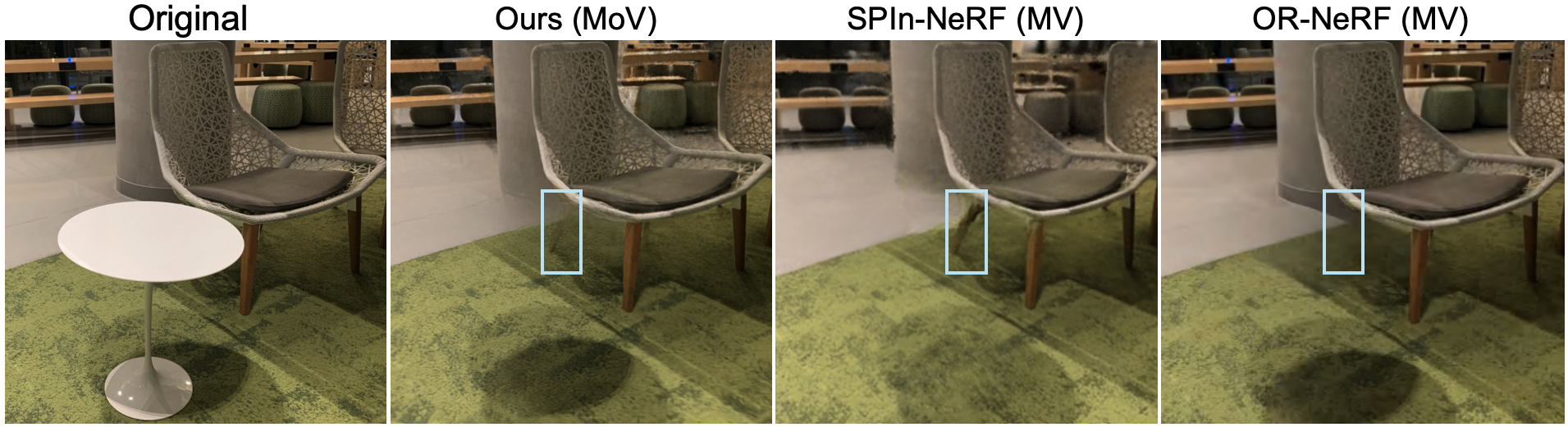}
    \vspace{-6pt}
    \caption{\textbf{Qualitative results on the IBRNet dataset.} In this example, we seek to inpaint the obstructed non-target object (highlighted box region). Herein, `MoV' refers to mono-view, while `MV' represents multi-view.}
    \label{fig:ibr-sota}
\end{figure}

The reasons behind these significant performance gains are further discussed in the ablation study (\cref{subsec:abl}). We provide qualitative comparisons with other baselines to investigate the effectiveness of our method in eliminating artifacts after inpainting. As illustrated in the second row in \cref{fig:vis spin sota}, other baselines exhibit noticeable black artifacts in the region where the kettle is removed, whereas our method produces a much smaller shadow on the chair. Additional visualization results can be found in the supplementary material. In addition, we employ IBRNet data for more qualitative results, which are shown in \cref{fig:ibr-sota}. Compared with the incomplete synthesis results from OR-NeRF, our method is still able to synthesize partial chair legs. Notice that SPIn-NeRF utilizes visible parts of chair legs from other views to complete their occluded regions, and thereby, their synthesis results are superior.

\subsection{Ablation Study} \label{subsec:abl}
\begin{figure}[t]
    \centering
    \includegraphics[width=1.0\linewidth]{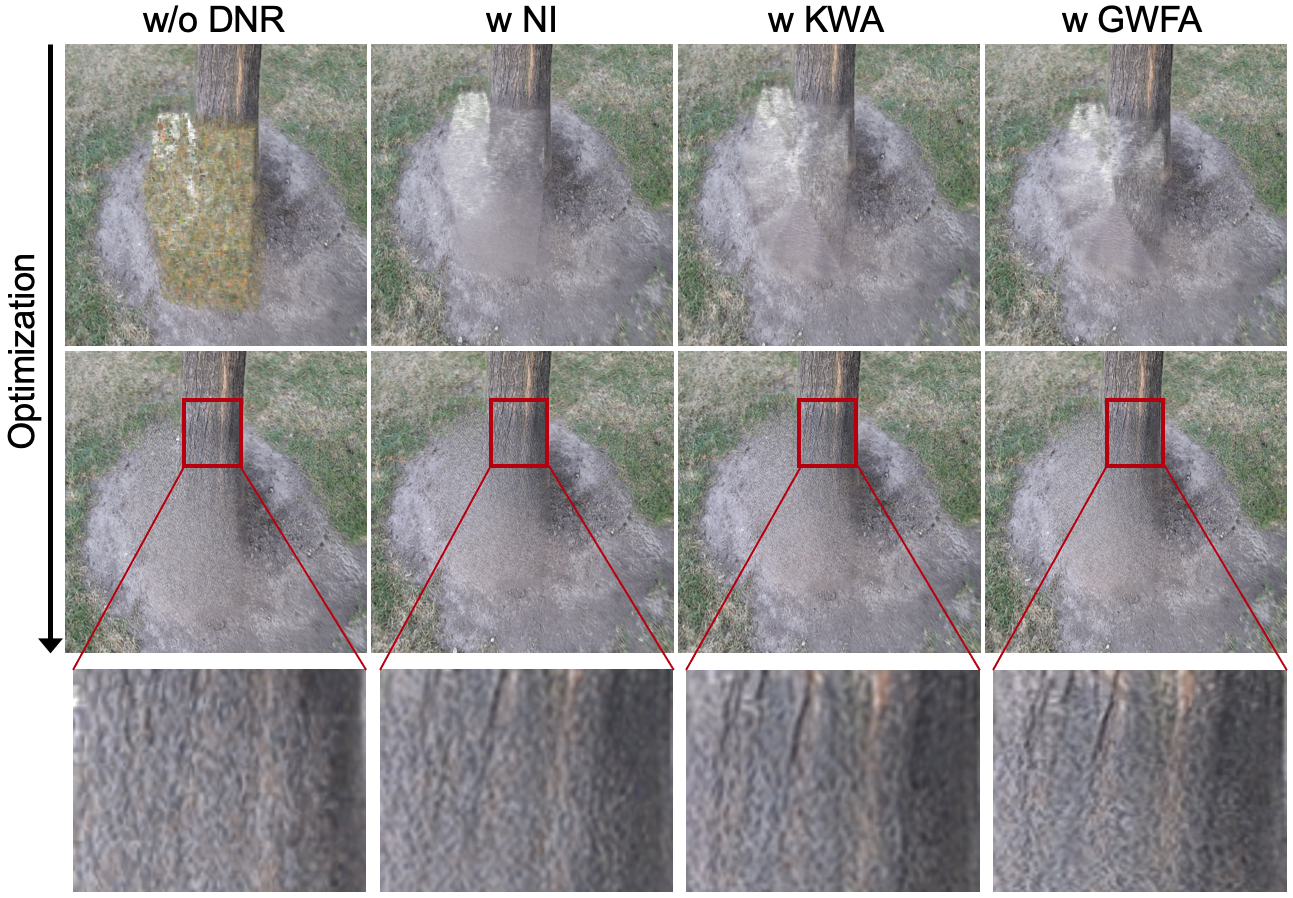}
    \caption{\textbf{Qualitative results implemented with different DNR strategies} on Scene 2 (SPIn-NeRF data \cite{mirzaei2023spin}). Left column: a comparison of the inpainted masked regions; Right column: the disparities among optimized tree textures. The red frames provide a detailed examination of the intricate texture.}
    \label{fig:dnr-vis}
\end{figure}

\minisection{Comparison Results of DNR Strategies.}
To validate the effectiveness of three DNR strategies, we conduct an ablation study on three scenes from SPIn-NeRF data. Average results are reported in \cref{tab:abl dnr}, and qualitative results after training for 3K steps are compared in \cref{fig:dnr-vis}.
\begin{table*}[htbp]
\begin{center}
\caption{\textbf{Comparison results subject to DNR strategies.} Row: Baseline (None) and methods with DNR strategies; Column: Training steps. Herein, c.s. denotes the convergence steps. For each set, we present LPIPS ($10^{-2}$)$\downarrow$ / FID$\downarrow$.}
\label{tab:abl dnr}
\tabcolsep=0.32cm
\adjustbox{width=1.0\linewidth}{
    \begin{tabular}{l|c|c|c|c|r}
    \hline
     & 0.5K & 1K & 2K & 3K & c.s.\\ 
    \hline
    None & 1.99 / 173.18 & 1.93 / 181.16 & 1.89 / 172.10 & 1.94 / 178.96 & 2.2K \\
    NI & 1.84 / 181.06 & 1.91 / 187.19 & 1.87 / 169.59 & 1.88 / 173.10 & 1.28K \\
    KWA & 1.84 / 184.36 & 1.86 / 189.36 & 1.84 / 176.55 & 1.88 / 184.52 & 1.2K \\
    GWFA & 1.82 / 187.70 & 1.84 / 190.57 & 1.86 / 183.39 & 1.86 / 183.08 & 1.16K \\
    \hline
    \end{tabular}}
\end{center}
\end{table*}
We observe that as the number of iterations increases, GWFA consistently outperforms the other two strategies, incidating that the introduction of GWFA aids to better results and faster convergence. We further time the average convergence in steps to enforce the similar level of training loss across three scenes, as reported in the last row of the table. It is noteworthy that NI exhibits the performance comparable to KWA in step 3K, but consistently outperforms it during steps from 0.5K to 2K. From this perspective, GWFA converges around step 1K and emerges as the fastest DNR strategy. Examining \cref{fig:dnr-vis}, a conspicuous observation is the evident improvement in the tree texture reconstruction, with GWFA demonstrating the most notable gains in recovering intricate texture details. This insight can also be concluded from \cref{fig:vis spin sota}.

\minisection{Comparison of Mutual Information in Inpainting.}
To validate the consistency between the theoretical and experimental findings, as well as visualization effects, we conduct an ablation experiment on mutual information (MI), as illustrated in \cref{tab:abl mi}. By varying from the formulations presented in \cref{subsec:infor-entro}, we calculate the average MI across all pairs of rays. For each scene, it is evident that the DNR significantly enhances MI, with GWFA consistently achieving the highest average. 
We observe that the MI evolves in line with the performance in \cref{fig:dnr-vis}.
\begin{table}[t]
\begin{center}
\caption{\textbf{Comparison of mutual information (MI) on SPIn-NeRF dataset scenes over training steps.} ``None'' denotes the inpainting without DNR. Mutual information is averaged among all pairs of adjacent frames. In general, MI has been significantly increased with incorporating investigate of DNR strategies.}
\vspace{0.25cm}
\label{tab:abl mi}
\tabcolsep=0.3cm
\adjustbox{width=0.7\linewidth}{
    \begin{tabular}{l|c|c|c|c}
    \hline
    $\times 10^{-2}$ & Scene Book & Scene 2 & Scene Trash & Average \\ 
    \hline
    None & 24.41& 1.07& -17.69& 2.59\\
    NI & 35.84& 32.15& 12.00& 26.66\\
    KWA & 36.68& 35.19& 14.82& 28.89\\
    GWFA & 38.05& 35.89& 14.28& 29.40\\
    \hline
    \end{tabular}
    }
\end{center}
\end{table}

\begin{figure}[t]
    \centering
    \includegraphics[width=1.0\linewidth]{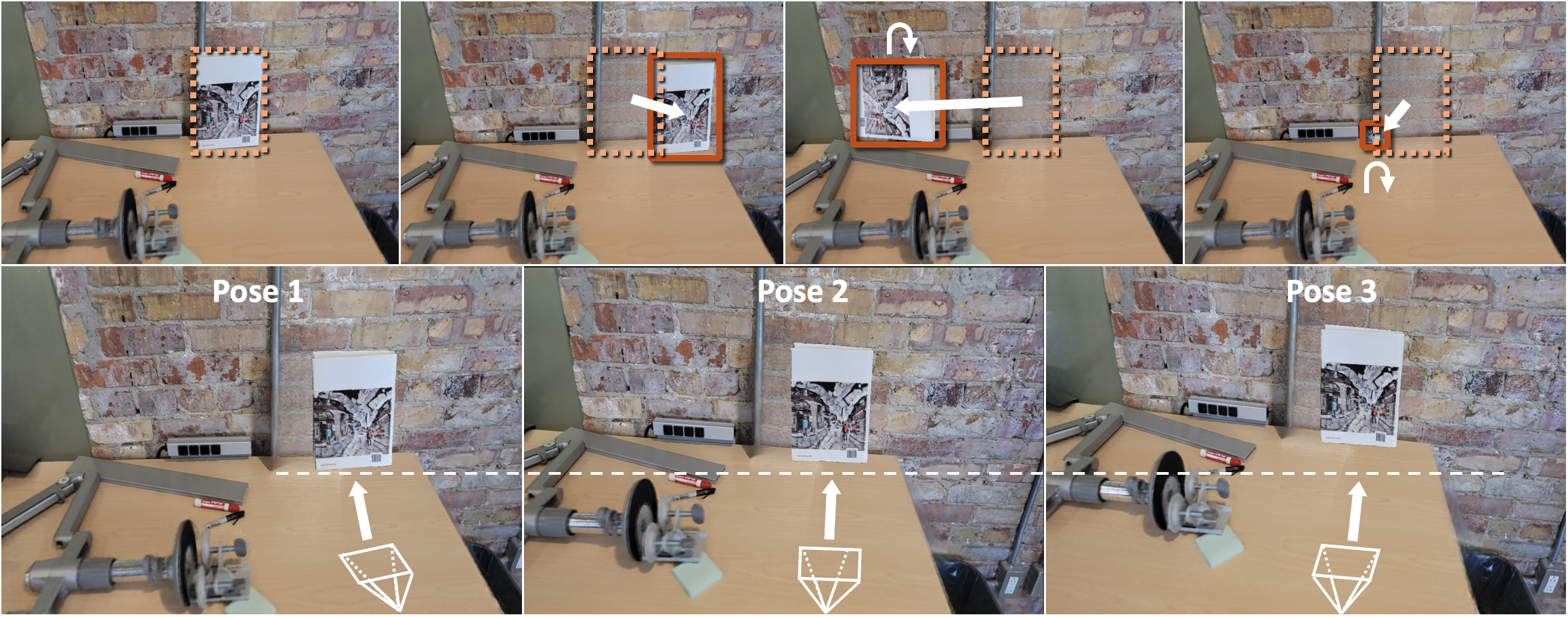}
    \caption{\textbf{Visualization of general ray transformations.} Top: Original, Translation, Rotation, Scaling. The orange and red frames refer to the object poses before and after transformation. Bottom: Rendered novel views. Validating rendering consistency with ray transformation and DNR inpainting. More results in \textbf{Supplementary}.}
    \label{fig:gen-trans}
\end{figure}

\minisection{Qualitative Results on General Transformation.}
As depicted in \cref{fig:gen-trans}, we showcase rendering outcomes corresponding to the transformation of a target object, involving operations such as rotation, translation, and scaling. 
We seek to conduct a preliminary exploration of the effects of editing operations on neural rays on the 3D locations and morphology of the target object.
We observe that the rendering quality in the original location of the target object is effectively inpainted, underscoring the success of ray operations in the broader context of general scene editing tasks, encompassing both rigid and non-rigid transformations. 
The second row illustrates the consistency of object transformation to different viewpoints. It can be observed that our ray manipulation and DNR exhibit good consistency in background completion across rendered new perspectives. Additional rendering results can be found in the supplementary materials, while video results are available on the \href{https://sample-nerf.github.io/}{Demos}.

\section{Conclusion}
This work contributes three advancements to object removal \& scene inpainting tasks within the research field of scene editing. First of all, our approach allows for direct scene manipulation through implicit ray transformations and produces visually consistent outcomes, aiming to reduce the difficulties of generating supervisions in object editing tasks. Then, we analyze the inpainting process from an informative standpoint and reveal that feature aggregation can enhance mutual information (MI) among rays, boosting overall performance. Consequently, we propose the novel Differentiable Neural-Point Resampling (DNR) to inpaint empty regions after editing. Ultimately, we validate the effectiveness of the ray transformation and DNR strategies. Our \model has achieved state-of-the-art performance on the removal \& inpainting task.

\minisection{Limitations.} 
In our study, supervision primarily stems from pretrained models in two folds. The initialization of point clouds relies on the depth estimation model, while the object segmentation and empty background optimization depend on the quality of target object masks as well as the scene inpainting model.
While these factors may affect the overall performance given the diversity of scenes (example failure cases can be found in the supplementary material), tackling these outliers is orthogonal to the scope of this work. In future work, we plan to jointly optimize depth estimation with target object masks, and incorporate DNR into the NeRF rendering process. 


\newpage







\bibliographystyle{plain}
\bibliography{main}
\end{document}